\documentclass[runningheads]{llncs}

\usepackage{eccv}

\usepackage{eccvabbrv}
\usepackage[ruled,linesnumbered]{algorithm2e}
\usepackage{graphicx}
\usepackage{booktabs}

\usepackage[accsupp]{axessibility}  
\newcommand\blfootnote[1]{%
  \begingroup
  \renewcommand\thefootnote{}\footnote{#1}%
  \addtocounter{footnote}{-1}%
  \endgroup
}

\usepackage{hyperref}

\usepackage{orcidlink}

\begin{document}

\title{GVGEN: Text-to-3D Generation with Volumetric Representation}


\author{Xianglong He\inst{1,2*} \and
Junyi Chen\inst{1,3*} \and 
Sida Peng\inst{4} \and
Di Huang\inst{1} \and
Yangguang Li\inst{5} \and
Xiaoshui Huang\inst{1} \and
Chun Yuan\inst{2\dagger} \and
Wanli Ouyang\inst{1,6} \and
Tong He\inst{1\dagger}
}

\authorrunning{He et al.}

\institute{$^1$Shanghai AI Laboratory $^2$Tsinghua Shenzhen International Graduate School \\
$^3$Shanghai Jiao Tong University \quad $^4$Zhejiang University \\
$^5$VAST \quad $^6$The Chinese University of Hong Kong 
}

\maketitle

\begin{abstract}
  In recent years, 3D Gaussian splatting has emerged as a powerful technique for 3D reconstruction and generation, known for its fast and high-quality rendering capabilities.
  Nevertheless, these methods often come with limitations, either lacking the ability to produce diverse samples or requiring prolonged inference times. 
  To address these shortcomings, this paper introduces a novel diffusion-based framework, GVGEN, designed to efficiently generate 3D Gaussian representations from text input. 
  We propose two innovative techniques:
  (1) \textit{Structured Volumetric Representation}.
  We first arrange disorganized 3D Gaussian points as a structured form GaussianVolume. This transformation allows the capture of intricate texture details within a volume composed of a fixed number of Gaussians. To better optimize the representation of these details, we propose a unique pruning and densifying method named the Candidate Pool Strategy, enhancing detail fidelity through selective optimization.
  (2) \textit{Coarse-to-fine Generation Pipeline}.
  To simplify the generation of GaussianVolume and empower the model to generate instances with detailed 3D geometry, we propose a coarse-to-fine pipeline. It initially constructs a basic geometric structure, followed by the prediction of complete Gaussian attributes.
  Our framework, GVGEN, demonstrates superior performance in qualitative and quantitative assessments compared to existing 3D generation methods. Simultaneously, it maintains a fast generation speed ($\sim$7 seconds), effectively striking a balance between quality and efficiency. Our project page is \url{https://gvgen.github.io/}.
  \blfootnote{$*$ Equal Contribution.}
  \blfootnote{$\dagger$ Corresponding Authors.}
\end{abstract}

\section{Introduction}
\label{Introduction}

The development of 3D models is a pivotal task in computer graphics, garnering increased attention across various industries, including video game design, film production, and AR/VR technologies. 
Among the different aspects of 3D modeling, generating 3D models from text descriptions has emerged as a particularly intriguing area of research due to its accessibility and ease of use.
Various methods~\cite{jain2022zero, poole2022dreamfusion, nichol2022point} have been proposed to handle the task.
Still, it continues to present difficulties owing to the ambiguity of texts and intrinsic domain gap between text description and corresponding 3D assets.

Previous text-to-3D approaches can be broadly classified into two categories: optimization-based generation~\cite{poole2022dreamfusion, yu2023text, chen2023text, wang2024prolificdreamer} and feed-forward generation~\cite{nichol2022point, ntavelis2024autodecoding, tang2023volumediffusion, li2023instant3d, MIR-2022-12-378face}.
Optimization-based methods have become rather popular recently, due to the rapid development of text-to-image diffusion models~\cite{dhariwal2021diffusion, rombach2022high}. These methods usually optimize 3D objects conditioned on texts or images through Score Distillation Sampling (SDS) ~\cite{poole2022dreamfusion}, distilling rich knowledge from 2D image generation models. 
Despite yielding impressive results, optimization-based methods face the Janus problem~\cite{poole2022dreamfusion}, manifesting as multiple faces or over-saturation problems. 
Additionally, the optimization of a single object can be prohibitively time-consuming, requiring extensive computational effort.
Contrarily, feed-forward approaches strive to generate 3D assets directly from text descriptions, thus sidestepping the Janus problem and significantly hastening the generation process. Our work is closely related to feed-forward-based methods. 
However, feed-forward methods that utilize multi-view generation models often create lower-resolution 3D assets than their multi-view image counterparts. Moreover, models directly generating 3D objects from texts often encounter difficulties with semantics when complex prompts are used.

Different from previous feed-forward-based methods like~\cite{li2023instant3d} that follow a text-2D-3D framework, our method proposes to generate 3D assets via directly learning 3D representation. 
In this study, we introduce an innovative and streamlined coarse-to-fine generation pipeline, GVGEN, for generating 3D Gaussians directly from text descriptions. Leveraging the highly expressive and fast-rendering capabilities of 3D Gaussians, our method achieves not only promising results but also maintains rapid text-to-3D generation and rendering.
As shown in \cref{fig:pipeline}, our method consists of two stages: GaussianVolume fitting and text-to-3D generation. 
In the first stage, we introduce GaussianVolume, a structured volumetric form composed of 3D Gaussians. 
Achieving this is challenging due to the sparse and unstructured nature of optimizing original Gaussians. 
To address this, we introduce a novel Candidate Pool Strategy for pruning and densification. 
The approach allows for fitting high-quality volumetric representation of Gaussians, rather than unordered points, making the generation process more conducive for a diffusion-based framework, as is utilized in the following step.

Despite the GaussianVolume establishing a structured volumetric framework that integrates seamlessly with existing diffusion pipelines, the intrinsic complexity of rich features of 3D Gaussians presents significant challenges. Specifically, capturing the distribution of a vast amount of training data effectively becomes difficult, resulting in hard convergence for the diffusion model.
Addressing these challenges, we partition the text-to-3D generation into two steps: coarse geometry generation and Gaussian attributes prediction.
To be more specific, in the first step, we employ a diffusion model to generate the coarse geometry of objects, termed the Gaussian Distance Field (GDF) - an isotropic representation outlining the proximity of each grid point to the nearest Gaussian point's center.
Following this, the generated GDF, in conjunction with text inputs, is processed through a 3D U-Net-based model to predict the attributes of GaussianVolumes, ensuring enhanced control and model convergence.

To the best of our knowledge, this is the first study to directly feed-forward generate 3D Gaussians from texts, exploring new avenues for rapid 3D content creation and applications. 
Our main contributions are summarized as follows:

\begin{itemize}
\item We introduce GaussianVolume, a structured, volumetric form consisting of 3D Gaussians. Through the innovative Candidate Pool Strategy for pruning and cloning, we accommodate high-quality GaussianVolume fitting within a fixed volumetric resolution. This framework seamlessly integrates with existing generative networks, leveraging the inherent advantages of 3D Gaussians for explicit and efficient representation.
\item We propose GVGEN, an efficient text-to-3D coarse-to-fine generation pipeline that first generates geometry volume and then predicts detailed Gaussian attributes, better controlling diverse geometry and appearances of generated assets. GVGEN achieves a fast speed ($\sim$7 seconds) compared with baseline methods, effectively striking a balance between quality and efficiency.
\item Compared with existing baselines, GVGEN demonstrates competitive capabilities in both quantitative and qualitative aspects.
\end{itemize}

\section{Related Works}
\subsection{Text-to-3D Generation}
Generating 3D objects conditioned on texts has become a challenging yet prominent research area in recent years. Previous approaches~\cite{mohammad2022clip, jain2022zero} utilize CLIP~\cite{radford2021learning} as a prior for 3D asset optimization but lacked realism and fidelity. With the rise of text-to-image generative models~\cite{dhariwal2021diffusion, rombach2022high}, Dreamfusion~\cite{poole2022dreamfusion} leverages Score Distillation Sampling (SDS) to generate diverse 3D objects, drawing on the rich prior knowledge embedded in these models.
Subsequently, some works~\cite{chen2023fantasia3d, li2023sweetdreamer, long2023wonder3d, liu2023unidream} focus on modeling and learning multi-modal attributes of objects (\rm{e.g.} colors, albedo, normals, and depths), to enhance consistency. ProlificDreamer~\cite{wang2024prolificdreamer} and LucidDreamer~\cite{EnVision2023luciddreamer} introduce novel losses to improve the quality.
Some studies~\cite{huang2023epidiff, shi2023zero123++, shi2023mvdream, liu2023syncdreamer, MIR-2023-06-083view} explore predicting multi-views for objects, followed by using feed-forward methods or SDS-based optimization to generate 3D objects. Moreover, integrating the recently proposed 3D Gaussian Splatting~\cite{kerbl20233d, chen2024survey}, works~\cite{tang2023dreamgaussian, yi2023gaussiandreamer, chen2023text} lead to improvements in convergence speed for text-to-3D object generation. However, these methods still require significant generation time ($\sim$hours) per object or face 3D inconsistency of generated objects.

As the emergence of large-scale 3D datasets~\cite{deitke2023objaverse, deitke2024objaverse}, Feed-forward models are trained to reduce the generation time. Point-E~\cite{nichol2022point} and Shap-E~\cite{jun2023shap}, trained on millions of 3D assets, generate point clouds and neural radiance fields respectively. 3D VADER~\cite{ntavelis2024autodecoding} trains auto-decoders to fit latent volumes and uses them to train the diffusion model. Similarly, VolumeDiffusion~\cite{tang2023volumediffusion} trains an efficient volumetric encoder to produce training data. Despite these advancements, text-to-3D Gaussian generation remains largely unexplored, primarily due to the complexity of organizing disorganized Gaussians.
Our work introduces a coarse-to-fine pipeline to feed-forward generate 3D Gaussians from texts, by first generating the object's coarse geometry and then predicting its explicit attributes, pioneering in the direct generation of 3D Gaussians from text descriptions.

\begin{figure}[!t]
\begin{center}
\centerline{\includegraphics[width=\linewidth]{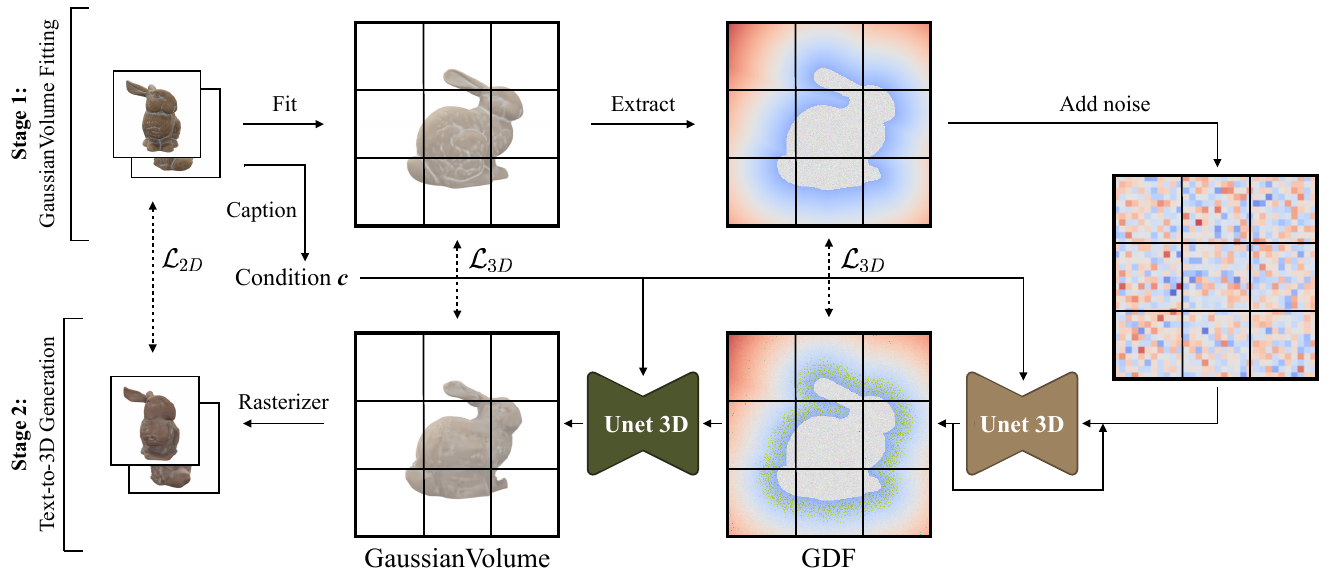}}
\caption{\textbf{Overview of GVGEN.} Our framework comprises two stages. In the data pre-processing phase, we fit GaussianVolumes (\cref{sec:3dgvf}) and extract coarse geometry Gaussian Distance Field (GDF) as training data. For the generation stage (\cref{sec:generation}), we first generate GDF via a diffusion model, and then send it into a 3D U-Net to predict attributes of GaussianVolumes.}
\label{fig:pipeline}
\end{center}
\end{figure}

\subsection{Differentiable 3D Representation}
Since the proposal of Neural Radiance Field~\cite{mildenhall2021nerf}, various differentiable neural rendering methods~\cite{barron2021mip, muller2022instant, chen2022tensorf, cao2023hexplane} have emerged, demonstrating remarkable capabilities in scene reconstruction, novel view synthesis, and 3D generation tasks. Instant-NGP~\cite{muller2022instant} employs feature volumes for accelerations by querying features only at the corresponding spatial positions. Works~\cite{chen2022tensorf, cao2023hexplane} decompose features into lower dimensions for faster training and less storage. However, they still have some shortcomings compared with point-based rendering methods~\cite{xu2022point, chang2023pointersect} in the aspects of rendering speed and explicit manipulability.

In recent research, 3D Gaussian Splatting~\cite{kerbl20233d} has received widespread attention. It adopts anisotropic Gaussians to represent scenes, achieving real-time rendering and facilitating downstream tasks like 3D generation~\cite{xu2024agg, ren2023dreamgaussian4d, yin20234dgen}, scene editing~\cite{chen2023gaussianeditor} and dynamic scene rendering~\cite{wu20234d}. We introduce a strategy to fit 3D Gaussians as a structured volumetric form called GaussianVolume, which allows for good integration with the existing generation pipeline, leveraging the advantages of 3D Gaussians and achieving fast generation and rendering speeds.

\section{Methodology}
\label{sec:method}

Our text-to-3D generation framework (\cref{fig:pipeline}), GVGEN, is delineated into two pivotal stages: GaussianVolume fitting (\cref{sec:3dgvf}) and text-to-3D generation (\cref{sec:generation}). 
Initially, in the GaussianVolume fitting stage, we propose a structured, volumetric form of 3D Gaussians, termed GaussianVolume. We fit the GaussianVolume as training data for the generation stage. This stage is crucial for arranging disorganized point-cloud-like Gaussian points as a format more amenable to neural network processing. 
To address this, We use a fixed number (\rm{i.e.} fixed volume resolution for the volume) of 3D Gaussians to form our GaussianVolume (\cref{sec:3dgvf}), thereby facilitating ease of processing. Furthermore, we introduce the Candidate Pool Strategy (CPS) for dynamic pruning, cloning, and splitting of Gaussian points to enhance the fidelity of fitted assets. Our GaussianVolume maintains high rendering quality with only a small number (32,768 points) of Gaussians.

In the phase of generation \cref{sec:generation}, we first use a diffusion model conditioned on input texts to generate the coarse geometry volume, termed Gaussian Distance Field (GDF), which represents the geometry of generated objects. Subsequently, a 3D U-Net-based reconstruction model utilizes the GDF and text inputs to predict the attributes of the final GaussianVolume, thus achieving the generation of detailed 3D objects from text descriptions.

\subsection{Stage1: GaussianVolume Fitting}
\label{sec:3dgvf}

\begin{figure}[!t]
    \begin{center}
        \centerline{\includegraphics[width=\linewidth]{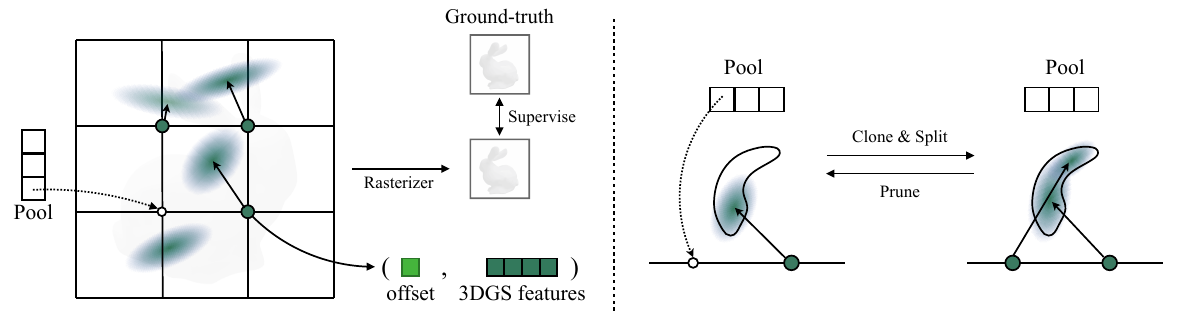}}
        \caption{\textbf{Illustration of GaussianVolume Fitting.} We organize a fixed number of 3D Gaussians in a volumetric form, termed GaussianVolume. By using position offsets to express slight movements from grid points to Gaussian centers, we can capture the details of objects. The proposed Candidate Pool Strategy (CPS) (\cref{sec:3dgvf}) enables effective pruning and densification with a pool storing pruned points.}
        \label{fig:fitting}
    \end{center}
\end{figure}

The original 3D Gaussian Splatting offers fast optimization and real-time rendering. However, its lack of structure makes it challenging for 3D neural networks to process effectively. Directly generating 3D Gaussians through networks involves integrating existing point cloud networks, treating Gaussians as carriers of semantically rich features. However, previous work~\cite{wu2023sketch} has pointed out the difficulty of directly generating point clouds with only 3-dimensional color features. Extending such methods, especially for high-dimensional Gaussians, may make the learning process very challenging. We also attempted to extend previous work~\cite{melas2023pc2} using point cloud diffusion to generate 3D Gaussians, while the results were disappointing.

Recent works~\cite{zou2023triplane, xu2024agg} propose hybrid representations combining 3D Gaussians with triplane-NeRF, utilizing point clouds for geometry and triplane features for texture querying, to finally decode 3D Gaussians. While these methods demonstrate effectiveness, the reconstructed results are heavily constrained by the accuracy of predicted point clouds and lose the direct operability advantage of 3D Gaussians. In light of these challenges, we introduce GaussianVolume, which is a volume composed of a fixed number of 3D Gaussians. This innovation facilitates the processing of 3D Gaussians by existing generation and reconstruction models, retaining the efficiency benefits of Gaussians. The illustration of the fitting process can be found in \cref{fig:fitting}.

\begin{algorithm}[tb]
   \caption{GaussianVolume Fitting with Candidate Pool Strategy}
   \label{alg:one}
   \KwIn{The number of Gaussian points $N$ with assigned positions $p$, total iteration times $T$, multi-view images ${I}$ with camera poses ${V}$, prune and densify conditions $\tau_p, \tau_d$}
   \KwOut{GaussianVolume $G$, a volumetric form composed of a fixed number of 3D Gaussians}
   $\ t\gets 0$; \tcp{Iteration Time}
   $G\gets$ Initialize($N, p$); \tcp{Initialize Gaussian Points}
   $P\gets \emptyset$; \tcp{Initialize Candidate Pool}
   \While{$t < T$}
       {$\ \hat{I}$ $\gets$ Rasterize($G, V$); \tcp{Rendering}
       $\mathcal{L}$ $\gets$ Loss($I$, $\hat{I}$); \tcp{Loss}
       $G$ $\gets$ Optimize($\nabla \mathcal{L}$); \tcp{Update Gaussian Attributes}
       \If{IsRefinementIteration($t$)}
            {$G_p\gets$ PrunePoints($G, \tau_p$);
               \tcp{Determine Pruned Points}
            $G, P \gets$ AddPointsToPool($G_p, G, P$); 
               \tcp{Make $G_p$ "Deactivated"}
            
            $G_d\gets$ DensifyPoints($G, \tau_d$);
               \tcp{Determine Densified Points in $G$}
            $G_{new} \gets$ FindPointsInPool($G_d, P$);
               \tcp{Find Added Points in $P$}
            $G, P \gets$ RemovePointsFromPool($G_{new}, G, P$) ;
               \tcp{Make $G_{new}$ "Activated"}
            }
       \If{EndRefinementIteration($t$)}
            { ReleasePool($G, P$); \tcp{Make All Points "Activated"}
            }
       }
\end{algorithm}

\subsubsection{GaussianVolume}
\label{sec:3dgv}
In 3D Gaussian Splatting~\cite{kerbl20233d}, features of 3D Gaussians $G$ include: a center position $\mu \in \mathbb{R}^3$, covariance matrix $\Sigma$, color information $c \in \mathbb{R}^3$ (when SH order=0) and opacity $\alpha \in \mathbb{R}$. To better optimize the covariance matrix, $\Sigma$ is analogous to describing the configuration of an ellipsoid via a scaling matrix $S \in \mathbb{R}^3$ and a rotation matrix $R \in \mathbb{R}^{3\times 3}$, satisfying: 
\begin{equation}
    \label{eq:eqsigma}
    \Sigma = R S S^T R^T.
\end{equation}
In the implementation, $S, R$ could be stored as a 3D vector $s\in \mathbb{R}^3$ and a quaternion $q\in \mathbb{R}^4$, respectively. With these differentiable features, 3D Gaussians can be easily projected to 2D splats and employ neural point-based $\alpha$-blending technique to render 2D-pixel colors. An efficient tile-based rasterizer is used for fast rendering and backpropagation.

In this work, we represent each object with a volume $V\in \mathbb{R}^{C\times N \times N\times N}$ composed of 3D Gaussians, which is trained from multi-view images of the object. Here, $C, N$ represent the number of feature channels and volume resolution respectively, and the number of 3D Gaussians is $N^3$. We follow the convention and rendering methods in original works, but use position offsets $\Delta \mu$ to express slight movements between the position $p$ of each grid point in the volume and the center $\mu$ of the Gaussian point it represents:
\begin{equation}
    \mu = p + \Delta \mu.
\end{equation}
During the fitting phase, we only apply the backward operation on the offsets $\Delta \mu$, which allows the expression of more fine-grained 3D assets and also imposes a restriction to form better structures. Due to the fixed number of Gaussians, we cannot directly apply the original strategy for densification control. Instead, we propose the Candidate Pool Strategy for effectively pruning and cloning.

\subsubsection{Candidate Pool Strategy}
\label{sec:cps}

We can not directly apply the original interleaved optimization/density strategy to move 3D Gaussians , due to the dynamical changes in the number of Gaussians not suitable for a fixed number of Gaussians. Densifying or pruning Gaussians freely is not allowed since Gaussians are bijective with assigned grid points. A naive way to avoid the problem is to only adjust the offsets $\Delta \mu$ relying on gradient back-propagation. Unfortunately, we experimentally found that the movement range of Gaussian centers becomes largely limited without an optimization/density strategy, which leads to lower quality and inaccurate geometry (see \cref{fig:ablation}(a)). For this reason, we propose a novel strategy (\cref{alg:one}) to densify and prune the fixed number of Gaussians. The key point to our strategy is storing pruned points in a candidate pool $P$ for later densification.

We initially align Gaussian centers, $\mu$, with assigned positions, $p$ and set offsets $\Delta\mu=0$ (Line 2 in \cref{alg:one}). The candidate pool $P$ is initialized as an empty set (Line 3). This pool stores "deactivated" points, refering to pruned points during optimization – they are not involved in the forward and backward process. We optimize each asset for $T$ iterations.
Following original 3DGS~\cite{kerbl20233d}, thresholds $\tau_p, \tau_d$ on view-space position gradients are used for pruning and densifying every fixed iterations (\textit{IsRefinementIteration(t)} in Line 8).

During the refining process, once Gaussian points $G_p$ are pruned with pre-defined threshold $\tau_p$, they are added into candidate pool $P$, making them "deactivated" for rendering and optimization (Lines 9-10).
For densification, we first determine points $G_d$ that should be densified in the "activated" set of Gaussians $G$. Then newly added points $G_{new}$ are selected from candidate pool $P$ via some criterion (e.g., we use the nearest point within a certain distance $\epsilon_{offsets}$ to the densified points $G_d$). The corresponding coordinate offsets for the added points are calculated. They become "activated" for both forward and backward processes and are subsequently removed from the candidate pool $P$ (Lines 11-13).
At the end of optimization/density, all points in the candidate pool are reintroduced into the optimization for further refinement (Lines 15-17).

This strategy ensures that Gaussian points can adaptively move during the optimization process, to represent more intricate object shapes. Simultaneously, the resulting structured volumetric form maintains its physical meaning, demonstrating a balance between adaptability and well-defined structure.

\subsubsection{Training Loss}
\label{sec:stage1sup} The final loss for supervision is the original loss used in 3D Gaussian Splatting adding a regularization loss:
\begin{equation}
    \label{eq:eqoffsets}
    \mathcal{L}_{offsets}=Mean(\mathrm{ReLU}(| \Delta \mu - \epsilon_{offsets}|)),
\end{equation}
where $\epsilon_{offsets}$ is a hyper-parameter, to restrict the center of 3D Gaussians not too far from their corresponding grid points,
\begin{equation}
    \label{eq:eq1}
    \mathcal{L}_{fitting}=\lambda_1 \mathcal{L}_1+\lambda_2 \mathcal{L}_{SSIM} + \lambda_3 \mathcal{L}_{offsets}
\end{equation}
After training, the points are sorted in spatial order according to volume coordinates as training data for the generation stage. Once the training is over, 2D images of target objects could be rendered at an ultra-fast speed since each GaussianVolume is rather lightweight. More implementation details can be found in the supplemental materials.

\begin{figure}[!t]
    \begin{center}
        \centerline{\includegraphics[width=\linewidth]{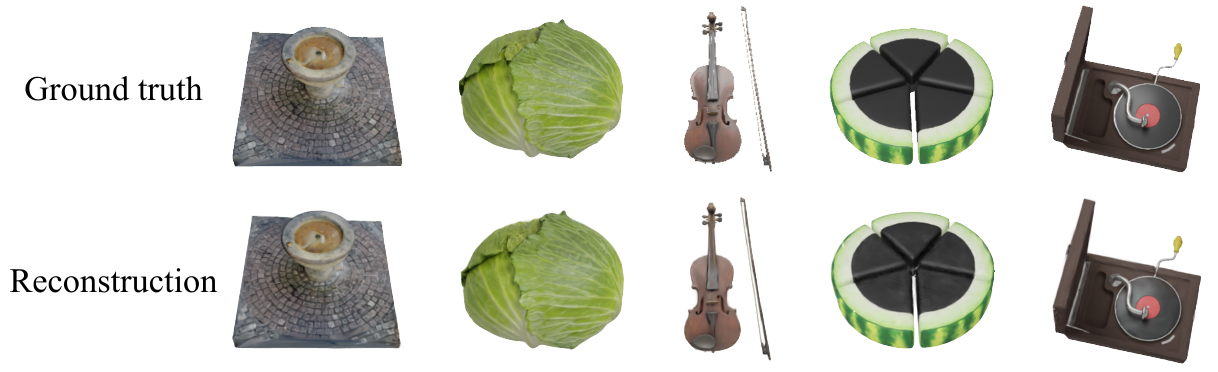}}
        \caption{\textbf{Visualization of GaussianVolume Fitting.} The rendering results demonstrate excellent reconstruction performance of GaussianVolumes.}
        \label{fig:stage1}
    \end{center}
\end{figure}

\subsection{Stage2: Text-to-3D Generation} 
\label{sec:generation}

As discussed in work~\cite{tang2023volumediffusion}, 3D volume space is high-dimensional, which potentially poses challenges in training the diffusion model. Empirically, we found that even though we form 3D Gaussians as a volumetric structure, it still takes too long for model convergence when training with a large amount of data.
Moreover, learning the data distribution of GaussianVolumes from diverse categories via a single diffusion model causes it to collapse to the single-step reconstruction, leading to the degradation of generation diversity. 
To overcome these challenges, we introduce a coarse-to-fine pipeline that first generates coarse geometry volumes (Gaussian Distance Field, GDF) and then predicts the attributes of the GaussianVolume. In the first step, we adopt a diffusion model conditioned on input texts to generate GDF. In the second step, we employ a 3D U-Net-based model to predict the attributes of Gaussians by inputting the GDF along with the text condition, leading to the final GaussianVolume.

\subsubsection{Gaussian Distance Field Generation}
\label{sec:gdf}

The Gaussian Distance Field (GDF) $F\in {\mathbb{R}^+_0}^{ 1\times N \times N\times N}$ stores an isotropic feature representing the fundamental geometry of a GaussianVolume. It measures the distance between each grid coordinate and its nearest Gaussian point, similar to the definition of Unsigned Distance Field. This attribute can be easily extracted from fitted GaussianVolumes via sorting algorithms.
After obtaining the ground truth GDF, we train a diffusion model conditioned on texts to generate GDF, creating the coarse geometry of objects. The model is supervised by minimizing Mean Square Error (MSE) loss $\mathcal{L}_{3D}$ between the ground truth GDF and the generated GDF, which is equivalent to predicting added noises in the diffusion process. Using diffusion models, the generation diversity with respect to object shape is introduced.

\subsubsection{GaussianVolume Prediction}
\label{sec:3dgp}
After generating GDF, we send it into a U-Net-based model modified from SDFusion~\cite{cheng2023sdfusion}, along with text descriptions, to predict all attributes of the GaussianVolume. The reason why a reconstruction model is adopted is that we empirically found that the single-step model produces comparable predictions of Gaussian attributes with a diffusion model.

We use two kinds of losses in this phase: MSE loss $\mathcal{L}_{3D}$ between ground truth GaussianVolume and the predicted one, and rendering loss $\mathcal{L}_{2D}$:
\begin{equation}
    \label{eq:eqrenderloss}
    \mathcal{L}_{2D}=\lambda \mathcal{L}_1+(1-\lambda)\mathcal{L}_{SSIM}
\end{equation}
for rendered images, which are composed of L1 and SSIM loss. The total loss is,
\begin{equation}
    \label{eq:eq2}
    \mathcal{L}=\lambda_{3D} \mathcal{L}_{3D}+\lambda_{2D} \mathcal{L}_{2D}
\end{equation}
Using a multi-modal loss balances global semantic and local details, and keeps the training process more stable. More implementation details about model architectures and data processing can be found in the supplemental materials.

\section{Experiments}

\begin{figure*}[tb]
    \begin{center}
        \centerline{\includegraphics[width=\linewidth]{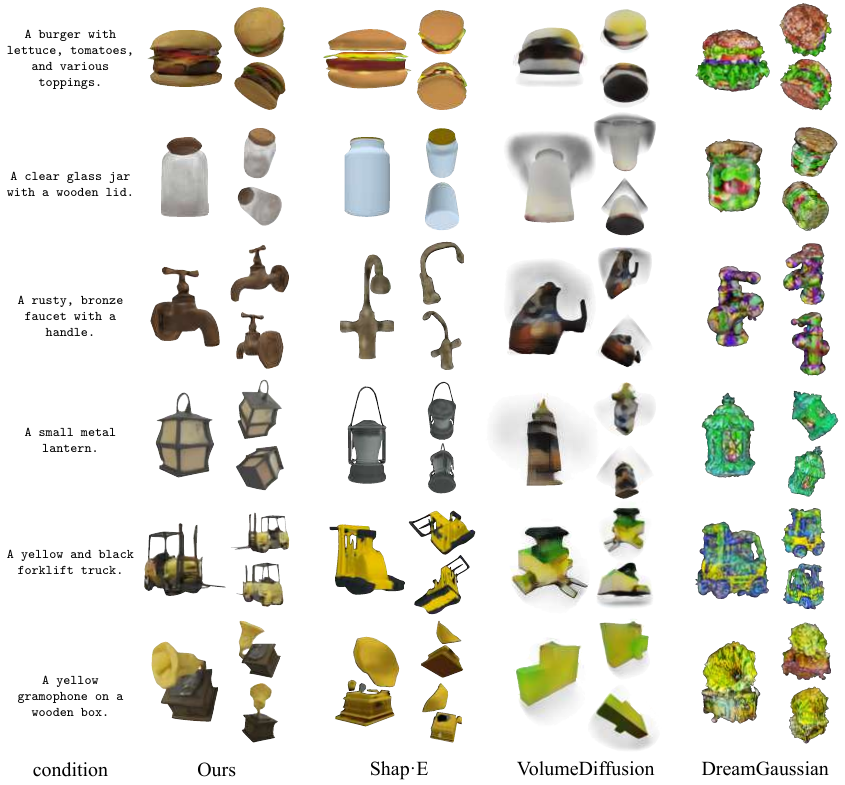}}
        \caption{\textbf{Comparisons with State-of-the-art Text-to-3D Methods.} Our method achieves competitive visual results with better alignments with text conditions.}
        \label{fig:display}
    \end{center}
\end{figure*}

\subsection{Baseline Methods and Dataset}
\subsubsection{Baseline Methods}
We compare with both feed-forward-based methods and optimization-based methods. Across the former methods, we evaluate Shap-E~\cite{jun2023shap} and VolumeDiffusion~\cite{tang2023volumediffusion}, both of which directly generate 3D assets. For latter ones, we consider DreamGaussian~\cite{tang2023dreamgaussian} as the baseline, where coarse 3D Gaussians undergo optimization with SDS loss~\cite{poole2022dreamfusion} and are converted into meshes for further refinement. For VolumeDiffusion, we report results generated in the feed-forward process, without post-optimization using SDS via pretrained text-to-image models, to ensure fair comparisons with other methods.

\subsubsection{Dataset}
Our training dataset comprises the Objaverse-LVIS dataset~\cite{deitke2023objaverse}, which contains $\sim$ 46,000 3D models in 1,156 categories. For text prompts for training, we use captions from Cap3D~\cite{luo2024scalable}, which leverages BLIP-2~\cite{li2023blip} to caption multi-view images of objects and consolidates them into single captions through GPT-4~\cite{achiam2023gpt}. For evaluation, we generate 100 assets and render 8 views for each asset. The camera poses for these views are uniformly sampled around the object.

\begin{figure}[tb]
    \begin{center}
        
        \centerline{\includegraphics[width=\linewidth]{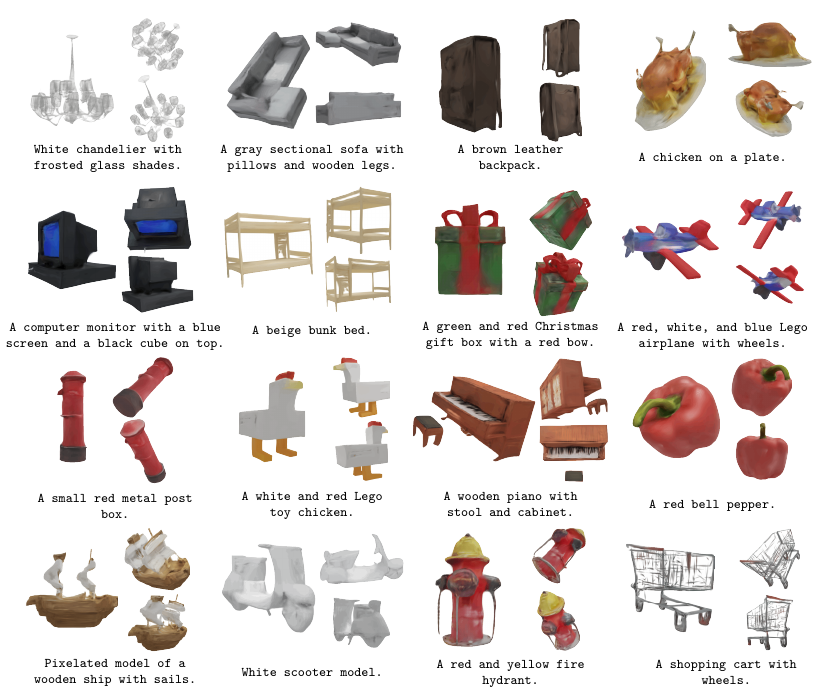}}
        \caption{\textbf{Text-to-3D Generation Results by GVGEN.}}
        \label{fig:gallery}
    \end{center}
\end{figure}

\subsection{Qualitative and Quantitative Results}
\subsubsection{GaussianVolume Fitting}
We present the reconstruction results of GaussianVolume Fitting in \cref{fig:stage1}. These results demonstrate that the proposed GaussianVolume can express high-quality 3D assets with a small number of Gaussian points under the volume resolution $N=32$. Higher resolutions yield better rendering effects but require more computational resources for the generation stage. Additional ablation studies about fitting GaussianVolume could be found in \cref{sec:ablation} and the supplemental materials.

\subsubsection{Text-to-3D Generation}
We provide visual comparisons and quantitative analyses of our method and existing baselines in \cref{fig:display} and \cref{tab:quantitative}, respectively. As depicted in the figure, our model generates reasonable geometry and plausible textures. Shap-E~\cite{jun2023shap} can produce rough shapes but sometimes result in misaligned appearances with respect to the input texts. VolumeDiffusion~\cite{tang2023volumediffusion} tends to generate unrealistic textures. And DreamGaussian~\cite{tang2023dreamgaussian}, which adopts an optimization-based approach, always produces over-saturated results. For quantitative results, we compare the CLIP score between rendered images and the corresponding texts, and also the generation time of these methods. We randomly select 300 text prompts for evaluation from the 40 LVIS categories containing the highest number of objects. For each object, we render 8 views in uniformly sampled camera poses. To measure generation time, we set the sampling steps to 100 for diffusion-based methods. The configuration of DreamGaussian~\cite{tang2023dreamgaussian} follows the default settings. Refer to \cref{fig:gallery} for more generation results.

\begin{figure}[!t]
\begin{center}

\centerline{\includegraphics[width=\linewidth]{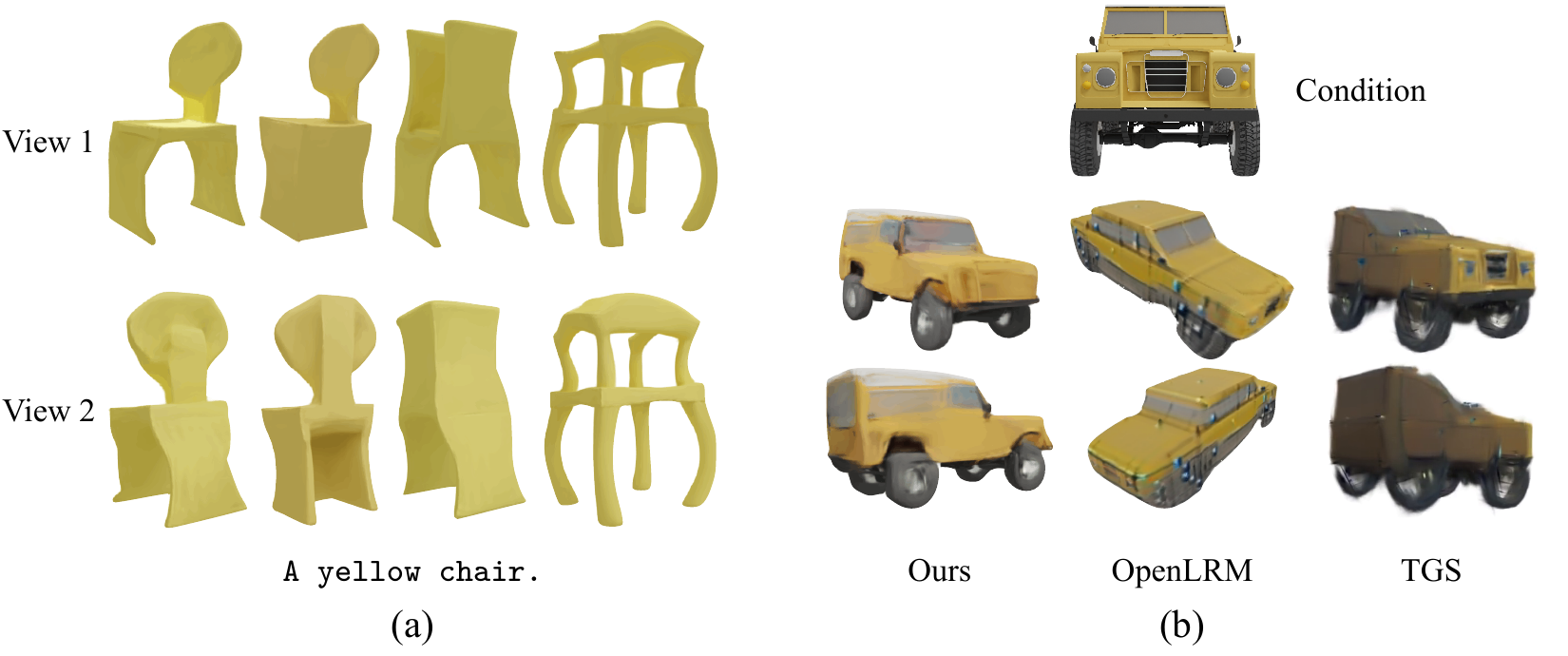}}
\caption{\textbf{Display of Generation Diversity.} (a) displays diverse results with the same text prompt via GVGEN. (b) shows the comparisons among our image-conditioned GVGEN, OpenLRM~\cite{openlrm} and TGS~\cite{zou2023triplane}. The single-view reconstruction models suffer from the problem of average patterns, leading to implausible shapes and textures in unseen regions, while GVGEN produces reasonable appearances and geometries.}
\label{fig:diversity}
\end{center}
\end{figure}

\subsubsection{Generation Diversity}
As depicted in \cref{fig:diversity}(a), GVGEN can generate diverse assets conditioned on the same prompt. The generative diversity of our method not only differentiates it from reconstruction approaches but also heightens the imaginative capabilities of users. Furthermore, we develop an image-to-3D model conditioned on CLIP image embeddings and compare the results with the recently popular single-image reconstruction models TGS~\cite{zou2023triplane} and LRM~\cite{hong2023lrm} (See \cref{fig:diversity}(b)). We use OpenLRM~\cite{openlrm} as an alternative to the close-sourced LRM. The single-view reconstruction models suffer from the problem of average patterns, leading to implausible shapes and textures in unseen regions, while GVGEN produces reasonable appearances and geometries. Such comparisons accentuate the critical difference between GVGEN and reconstruction methods.

\begin{table}[t]
\caption{\textbf{Quantitative Comparisons against Baseline Methods on CLIP Score and Inference Speed.}}
\label{tab:quantitative}
    \begin{center}
        \begin{tabular}{c|cc}
        
          \toprule
          Metrics        & CLIP score $\uparrow$ & Time $\downarrow$ \\
          \midrule
          
          Ours            & \bf{28.53}     & \bf{7 sec}   \\
          Shap-E          & 28.48     & 11 sec  \\
          VolumeDiffusion & 25.09     & 7 sec  \\
          DreamGaussian   & 23.60     & $\sim$3 min  \\
          \bottomrule
        \end{tabular}
    \end{center}
\end{table}

\begin{figure}[!t]
\begin{center}

\centerline{\includegraphics[width=\linewidth]{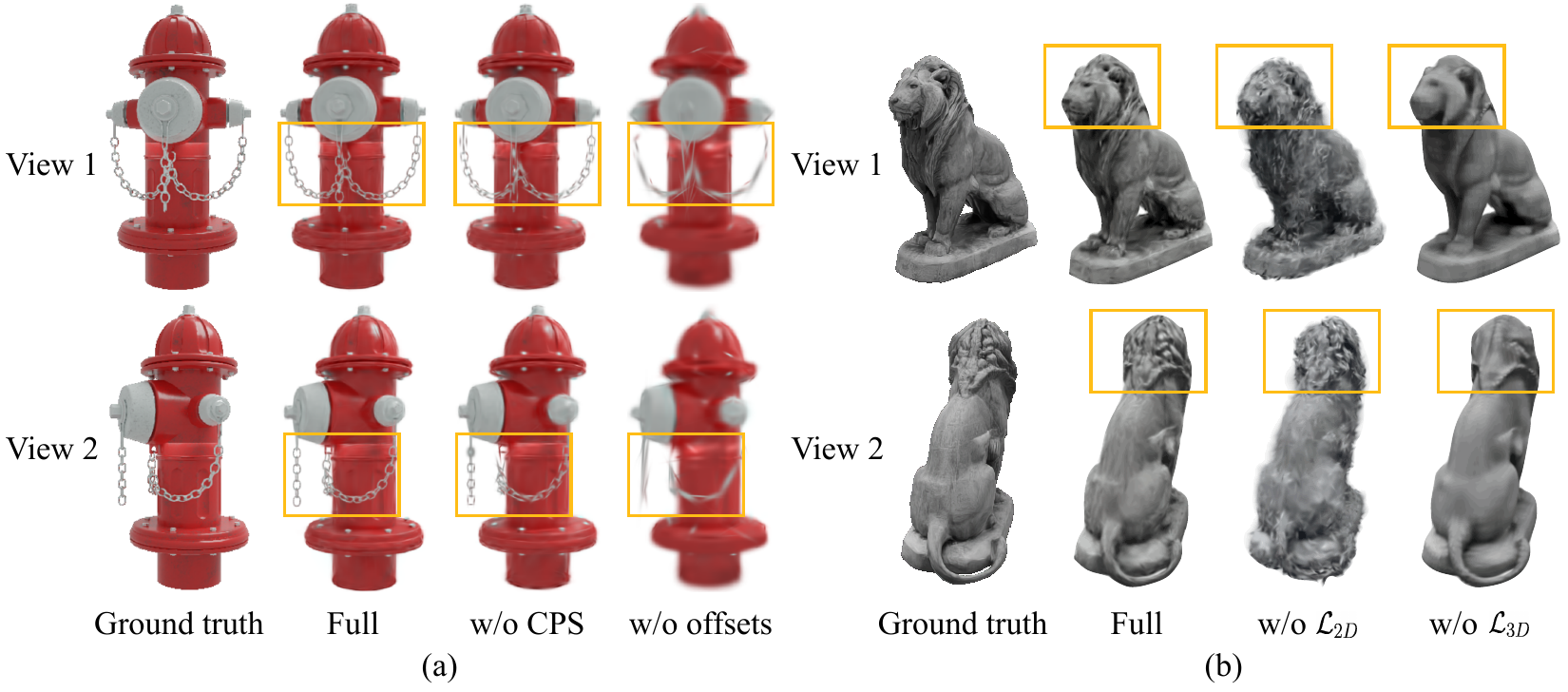}}
\caption{\textbf{Qualitative Results for Ablation Studies.} (a) represents visual comparisons of different GaussianVolume fitting methods. (b) stands for results using different losses to train the GaussianVolume prediction model.}
\label{fig:ablation}
\end{center}
\end{figure}

\begin{table}[t]
    \label{tab:ablation}
    \caption{\textbf{Quantitative Metrics for Ablation Studies.} The left table analyzes GaussianVolume fitting strategies, while the right table compares different losses training the prediction model. For qualitative comparisons, see \cref{fig:ablation}.}
    \begin{subtable}[h]{0.45\textwidth}
        \centering
        \begin{tabular}{c|ccc}

          \toprule
          Metrics        & PSNR $\uparrow$ & SSIM $\uparrow$ & LPIPS $\downarrow$ \\
          \midrule
          
          Full           & \bf{30.122}    & \bf{0.963}    & \bf{0.038}    \\
          w/o CPS    & 29.677    & 0.958    & 0.049    \\
          w/o offsets        & 27.140    & 0.936    & 0.084    \\
          \bottomrule
        \end{tabular}
        
        \label{tab:ablation1}
    \end{subtable}
    \hfill
    \begin{subtable}[h]{0.45\textwidth}
        \centering
        \begin{tabular}{c|ccc}

          \toprule
          Metrics        & PSNR $\uparrow$ & SSIM $\uparrow$ & LPIPS $\downarrow$ \\
          \midrule
          
          Full                         & 35.03    & \bf{0.9872} &  \bf{0.0236} \\
          w/o $\mathcal{L}_{3D}$       & \bf{35.21}    & 0.9846 &   0.0268 \\
          w/o $\mathcal{L}_{2D}$       & 29.55    & 0.9654 & 0.0444  \\
          \bottomrule
        \end{tabular}
        
        \label{tab:ablation2}
    \end{subtable}
     
     \label{tab:temps}
\end{table}

\subsection{Ablation Studies}
\label{sec:ablation}

\subsubsection{GaussianVolume Fitting}
We first study the effects of strategies fitting GaussianVolume via visual comparisons. In these ablation experiments, 3D assets are trained with 72 images, with camera poses uniformly distributed at a distance of 2.4 from the world center, and evaluated with 24 images posed uniformly at a distance of 1.6. We set volume resolution $N=32$, \rm{i.e.} number of Gaussian points $N^3=32,768$. The full method, as used in the data preparation stage, performs the best in terms of rendering results, even with only $32,768$ Gaussians. "w/o CPS" refers to optimizing coordinate offsets $\Delta \mu$ only via backpropagation, while "w/o offsets" means fixing Gaussian coordinates $\mu = p$ without optimizing point positions. \cref{fig:ablation}(a) shows that using the full method produces the best rendering results. \cref{tab:ablation1} reports the average PSNR, SSIM, and LPIPS scores, supporting the effectiveness of the proposed strategy quantitatively.

\subsubsection{Text-to-3D Generation}
Due to the computational resources required, it is impractical to evaluate each part of our design with models trained on the full Objaverse-LVIS dataset for our ablation study. Therefore, we train the models on a small subset of Objaverse-LVIS to validate the effectiveness of losses in the stage of predicting GaussianVolume attributes. Specifically, we pick 1,336 assets themed animals 
from 24 LVIS categories for training. 

We randomly select $\sim$20\% of the data from the training data and evaluate quantitative metrics with rendered novel views. \cref{fig:ablation}(b) demonstrates that the proposed multi-modal losses produce plausible results with fine-grained details, and quantitative results are listed in \cref{tab:ablation2}. The combination of two types of losses leads to more detailed textures while keeping the geometry smooth.

\subsection{Limitations}
GVGEN has shown encouraging results in generating 3D objects. However, its performance is constrained when dealing with input texts significantly divergent from the domain of training data. It is also time-consuming for fitting millions of training data to scale up for better diversity. Additionally, the volume resolution $N$ is set as 32 to save computational resources, which limits the rendering effects of 3D assets with very complex textures. In the future, we will further explore how to generate higher-quality 3D assets in more challenging scenarios.

\section{Conclusions}
In conclusion, this paper explores the feed-forward generation of explicit 3D Gaussians conditioned on texts. We innovatively organize disorganized 3D Gaussian points into a structured volumetric form GaussianVolume with a novel pruning and densifying strategy, \rm{i.e.} Candidate Pool Strategy, enabling feed-forward generation of 3D Gaussians via a coarse-to-fine generation pipeline. Our proposed framework, GVGEN, demonstrates remarkable efficiency in generating 3D Gaussians from texts. Experimental results demonstrate its competitive capabilities. This progress suggests potential extensions of our approach in tackling a broader spectrum of challenges within the field.

\subsubsection{Acknowledgments.} 
This work was done during Xianglong He's internship at Shanghai Artificial Intelligence Laboratory. The work is supported by the National Key R\&D Program of China (No. 2022YFB4701400/4701402, No. 2022ZD 0160102), SSTIC Grant (KJZD20230923115106012, KJZD20230923114916032), and Beijing Key Lab of Networked Multimedia. 

%
%
\bibliographystyle{splncs04}
\bibliography{main}
\end{document}